\documentclass{article}
\usepackage{spconf,amsmath,epsfig}
\usepackage{times}
\usepackage{graphicx}
\usepackage{amssymb}
\usepackage[breaklinks=true,letterpaper=true,colorlinks,bookmarks=false]{hyperref}
\usepackage[preprint]{spconfa4}

\def\ie{\emph{i.e.}}
\def\eg{\emph{e.g.}}
\def\etal{{\em et al.}}

\let\OLDthebibliography\thebibliography
\renewcommand\thebibliography[1]{
  \OLDthebibliography{#1}
  \setlength{\parskip}{0pt}
  \setlength{\itemsep}{0pt plus 0.3ex}
}

\begin{document}\sloppy

\def\x{{\mathbf x}}
\def\L{{\cal L}}

\title{Exploring Driving-aware Salient Object Detection via Knowledge Transfer}
%
\name{Jinming Su$^{1,3}$, Changqun Xia$^{2,*}$ and Jia Li$^{1,2,*}$}
\address{$^1$State Key Laboratory of Virtual Reality Technology and Systems, SCSE, Beihang University\\
$^2$Peng Cheng Laboratory, Shenzhen, China \quad $^3$Meituan \\ {\tt\small sujinming@meituan.com, xiachq@pcl.ac.cn, jiali@buaa.edu.cn}
}

\maketitle
\begin{abstract}
Recently, general salient object detection (SOD) has made great progress with the rapid development of deep neural networks. However, task-aware SOD has hardly been studied due to the lack of task-specific datasets.
In this paper, we construct a driving task-oriented dataset where pixel-level masks of salient objects have been annotated.
Comparing with general SOD datasets, we find that the cross-domain knowledge difference and task-specific scene gap are two main challenges to focus the salient objects when driving.
Inspired by these findings, we proposed a baseline model for the driving task-aware SOD via a knowledge transfer convolutional neural network.
In this network, we construct an attention-based knowledge transfer module to 
make up the knowledge difference. In addition, an efficient boundary-aware feature decoding module is introduced to perform fine feature decoding for objects in the complex task-specific scenes. The whole network integrates the knowledge transfer and feature decoding modules in a progressive manner. 
Experiments show that the proposed dataset is very challenging, and the proposed method outperforms 12 state-of-the-art methods on the dataset, which facilitates the development of task-aware SOD.
\end{abstract}
\let\thefootnote\relax\footnotetext{* Correspondence should be addressed to Changqun Xia and Jia Li. The dataset, code and models are available on \url{http://cvteam.net}.}

\section{Introduction}
\label{sec:intro}
Image-base salient object detection (SOD) aims to detect and segment objects that capture human visual attention, which is a preliminary step for subsequent vision tasks such as object recognition and tracking. 
Over the past decades, many benchmark datasets~\cite{yan2013hierarchical, yang2013saliency, wang2017stagewise} have been constructed for the study of SOD. These datasets include natural images with simple and complex scenes, which usually don't  focus on specific tasks. Such datasets can be referred to as general SOD datasets (or conventional SOD datasets). Based on these datasets, many SOD methods~\cite{zhang2017amulet,liu2018picanet,wang2018detect,su2019selectivity} have been proposed and achieve great performance. Correspondingly, this kind of task is called general SOD (or conventional SOD). However, in real world, what we usually need is to deal with specific application scenarios (\eg, automatic driving and medical image processing). For these scenarios, the analysis of salient objects plays an important role for assisting subsequent high-level visual tasks and understanding these scenes. As a distinction, this kind of task can be named as task-aware SOD. 

\begin{figure}[t]
\centering
\includegraphics[width=1\columnwidth,height=5cm]{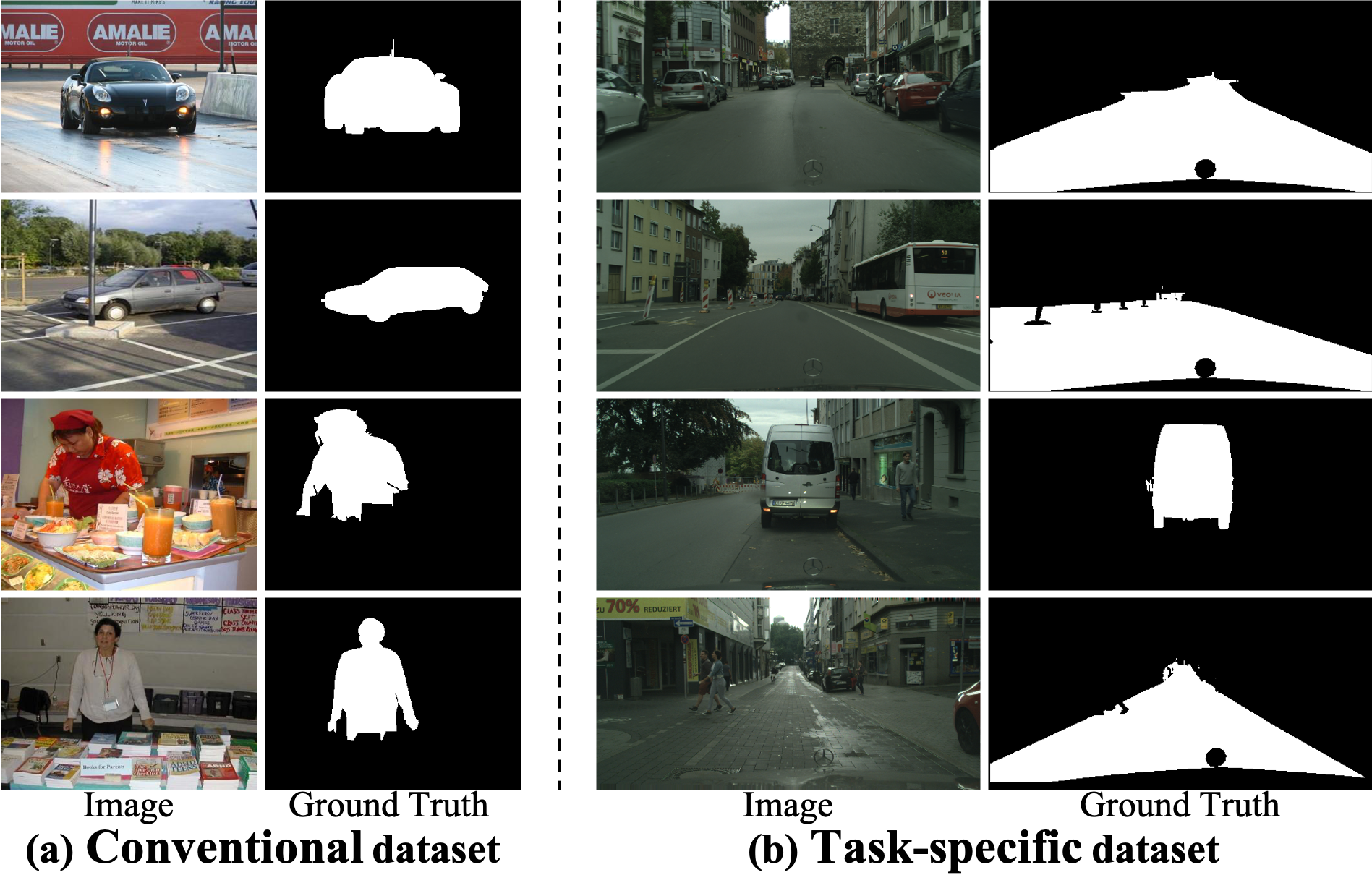}
\vspace{-7mm}
\caption{Comparisons of the conventional and task-specific SOD dataset. Images and ground-truth masks of the (a) conventional dataset and (b) task-specific dataset are from DUT-OMRON~\cite{yang2013saliency} and CitySaliency, respectively.}
\label{fig:dataset_motivation}
\end{figure}

Recently some task-specific datasets have been collected. These datasets contain images of specific tasks, such as driving, webpage-related behavior and game. For example, Alletto~\etal~\cite{alletto2016dr} proposed a publicly available actual driving dataset DR(eye)VE with task-specific fixation maps. Based on these datasets, task-aware fixation prediction (FP) has made progress. However, these still lack task-specific datasets for SOD, which handers the development of task-aware SOD.

Toward this end, we propose a new driving task-oriented dataset (denoted as \textbf{CitySaliency}), which contains 3,475 images with pixel-level annotation based on the common driving task. Some representative examples are shown in the task-specific dataset of Fig.~\ref{fig:dataset_motivation}. In constructing this dataset, we first collect these images from the driving scene. Given these images, we ask 25 volunteers to view these images by simulating 
\newpage
\pagestyle{empty}
\noindent
driving task to collect eye-tracking data and human fixations.
Based on these datas, we compute the saliency value of each object with the help of semantic annotations to determine ground-truth masks of salient objects. Finally, the images and masks together form the task-specific dataset. Comparing CitySaliency with conventional datasets, we find there exist two main challenges: 1) cross-domain knowledge difference. As shown in Fig.~\ref{fig:dataset_motivation}, cars  and persons in conventional datasets are the focus, while more attention in CitySaliency is paid to  drivable roads while ignoring vehicles (first two rows) and pedestrians (last two rows of Fig.~\ref{fig:dataset_motivation}) on both sides of roads. The knowledge difference about different task-related targets brings resistance to SOD. 2) task-specific scene gap. Based on the statistical analysis, we find that CitySaliency has three main characteristics including more discrete distribution, more salient objects and more area occupation, which reflects the complexity of scenes in CitySaliency.  These two problems make it difficult to deal with task-aware SOD.


To address these problems, we propose a baseline model for the driving task-aware SOD via a knowledge transfer convolutional neural network. In the network, we put forward an attention-based knowledge transfer module, which is used to transfer the knowledge from general domain to task domain to make up the knowledge difference, assisting the network to find the salient objects. In addition, we introduce an efficient boundary-aware feature decoding module to fine decode features for lots of existing salient objects with structures in complex task-specific scenes. The two modules are integrated in a progressive manner to deal with the task-aware SOD. Experiments show the proposed method outperforms 12 state-of-the-art methods on CitySaliency, which demonstrates the effectiveness of the proposed knowledge transfer network. 

The main contributions of this paper include: 1) we propose a new task-specific SOD dataset, which can be used to boost the development of task-aware SOD, 2) we propose a baseline model to deal with the difficulties in the proposed dataset and it consistently outperforms 12 state-of-the-art algorithms, 3) we provide a comprehensive benchmark of state-of-the-art methods and the proposed method on CitySaliency, which reveals the key challenges of the proposed dataset, and validates the usefulness of the proposed method.

\section{A New Dataset for Task-aware SOD}

We propose a new driving task-oriented dataset (named as \textbf{CitySaliency}) for task-aware SOD, which focuses on the common urban driving task. In this section, we will introduce the details in constructing the dataset.

\subsection{Data Collection}
The proposed dataset is built on Cityscapes~\cite{cordts2016cityscapes}, which is a complex urban driving dataset for instance-level semantic labeling. Cityscapes consists of 5,000 images with pixel-level semantic annotation (as shown in Fig.~\ref{fig:ddataset_annotation}(a)(b)), which is split into 2,975 image for training, 500 image for validating and 1,525 for testing (note the annotation of testing set is withheld for official benchmarking purposes). In constructing CitySaliency, we combine the training and validation sets (3,475 images) as a raw data source.

\subsection{Psychophysical Experiments}
To annotate salient objects in complex urban driving scenes, there exists a tough problem: there may exist several candidate objects while different annotator may have different biases in determining which objects are salient. To alleviate this influence, we conduct psychophysical experiments to collect human fixations as auxiliary information, as done in~\cite{li2017benchmark}. 

\begin{figure}[t]
\centering
\includegraphics[width=1.00\columnwidth,height=4.5cm]{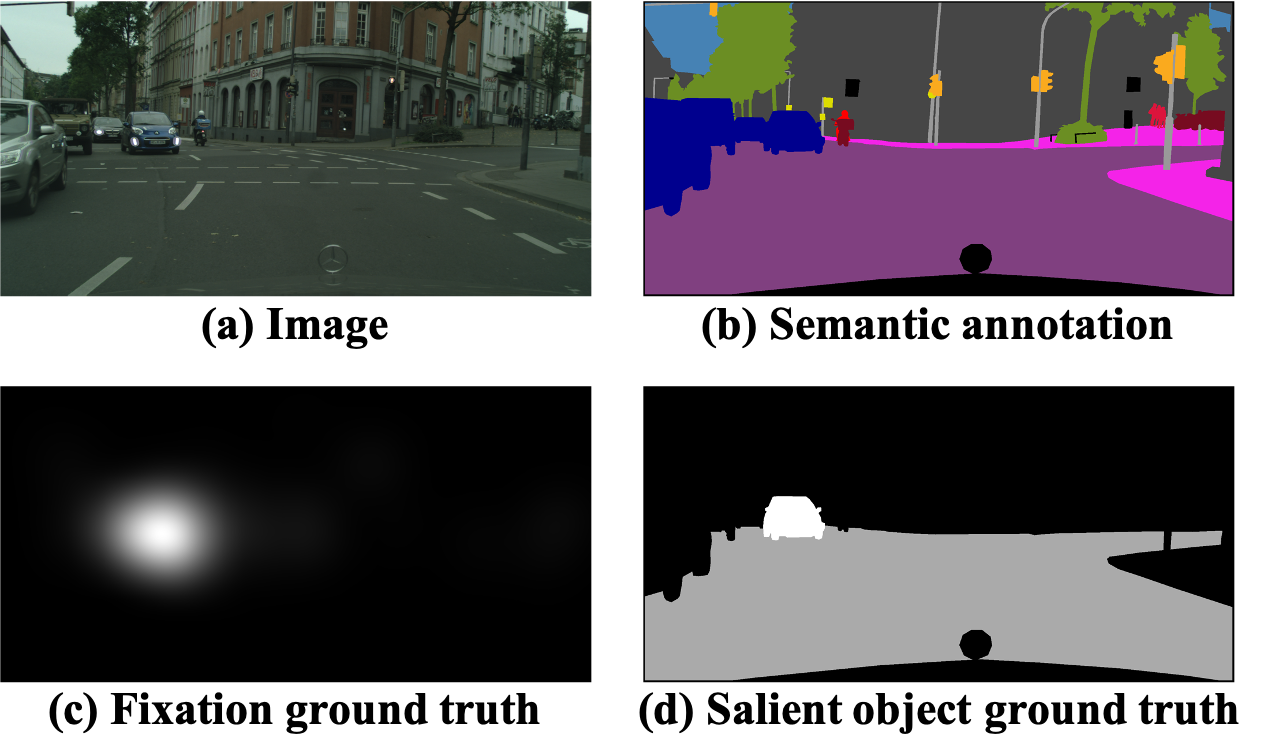}
\vspace{-7mm}
\caption{Process of annotating CitySaliency. For (a) an image, we get (c) the human fixations by psychophysical experiments. With (b) semantic annotation from Cityscapes~\cite{cordts2016cityscapes} and (c), we generate (d) the ground truth of salient objects.}
\label{fig:ddataset_annotation}
\end{figure}

In the psychophysical experiments, 25 subjects with normal vision and never seeing collected images participate in. During the experiment, we show the images on a 22-inch display screen with a resolution of 1680x1050 and install eye-tracking apparatus (SMI RED 500) with a sampling rate of 500HZ to record various eye movements including fixation, saccade and blink. 
After the data collection, various information of human eye movement is recorded. 
Based on the fixation data, we can compute the fixation density map for each image to annotate the ground-truth salient regions as in ~\cite{li2017benchmark}. 
The generated fixation ground truth as displayed in Fig.~\ref{fig:ddataset_annotation}(c).

\begin{figure}[t]
\centering
\vspace{-2mm}
\includegraphics[width=1.01\columnwidth,height=2.3cm]{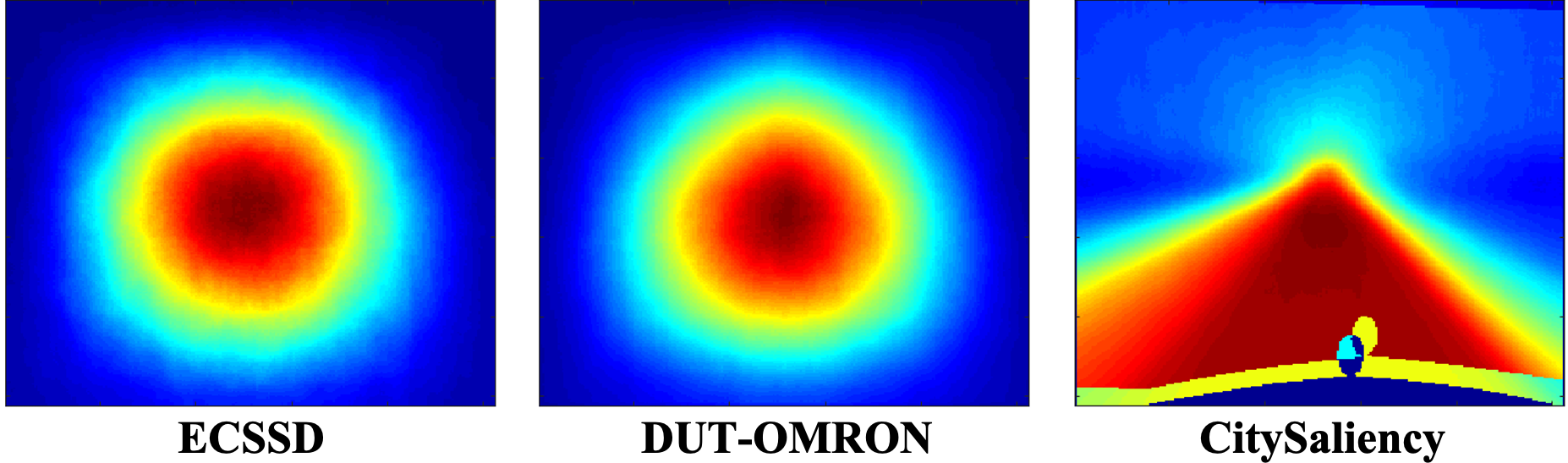}
\vspace{-7mm}
\caption{AAMs of two conventional datasets and CitySaliency.}
\label{fig:dataset_aam}
\end{figure}


\subsection{Generation of Salient Object Ground Truth}
Since Cityscapes provides pixel-level and instance-level semantic labeling, location of objects in CitySaliency is known. We represent the fixation density map of an image at time $t$ as $S_t$.
Based on the location of objects and the computed fixation density map, we can compute the saliency score $S(\mathcal{O})$ for each object $\mathcal{O} \in \mathcal{I}_t$ from the global perspective:
\begin{equation}
\begin{split}
& S(\mathcal{O}) = \\
& \frac{\sum_{\mathcal{I}_t \in \mathcal{V}} \text{I}(\mathcal{O} \in \mathcal{I}_t) \cdot \left( (1 + \frac{1}{\Vert\mathcal{O}\Vert}) \sum_{p \in \mathcal{O}} S_t(p) \right)}{\sum_{\mathcal{I}_t \in \mathcal{V}} \text{I}(\mathcal{O} \in \mathcal{I}_t)},
\end{split}
\label{eq:dataset_annotation}
\end{equation}
where $\Vert\mathcal{O}\Vert$ is the number of pixels in object $\mathcal{O}$. In Eq.~(\ref{eq:dataset_annotation}), saliency of an object is defined as the sum of its total fixation density and its average fixation density. After that, we select objects with saliency scores above an empirical threshold of top 80\% of the maximum saliency score. In this manner, we get several salient objects with pixel-wise instance-level annotation and different saliency scores as shown in Fig.~\ref{fig:ddataset_annotation}(d). 

\begin{figure}[t]
\centering
\includegraphics[width=0.90\columnwidth, height=2.7cm]{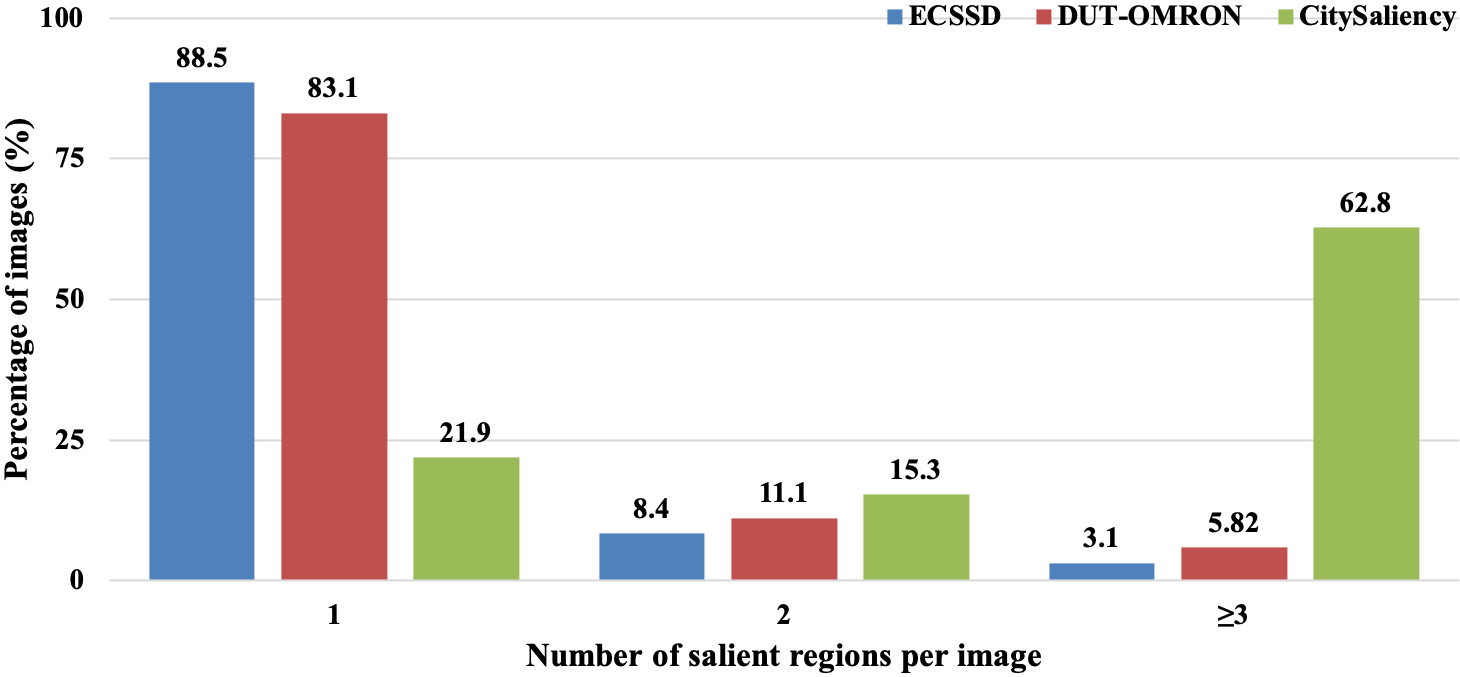}
\vspace{-4mm}
\caption{Histograms of the number of salient objects.}
\label{fig:dataset_number}
\end{figure}

\begin{figure}[t]
\centering
\includegraphics[width=0.90\columnwidth, height=2.7cm]{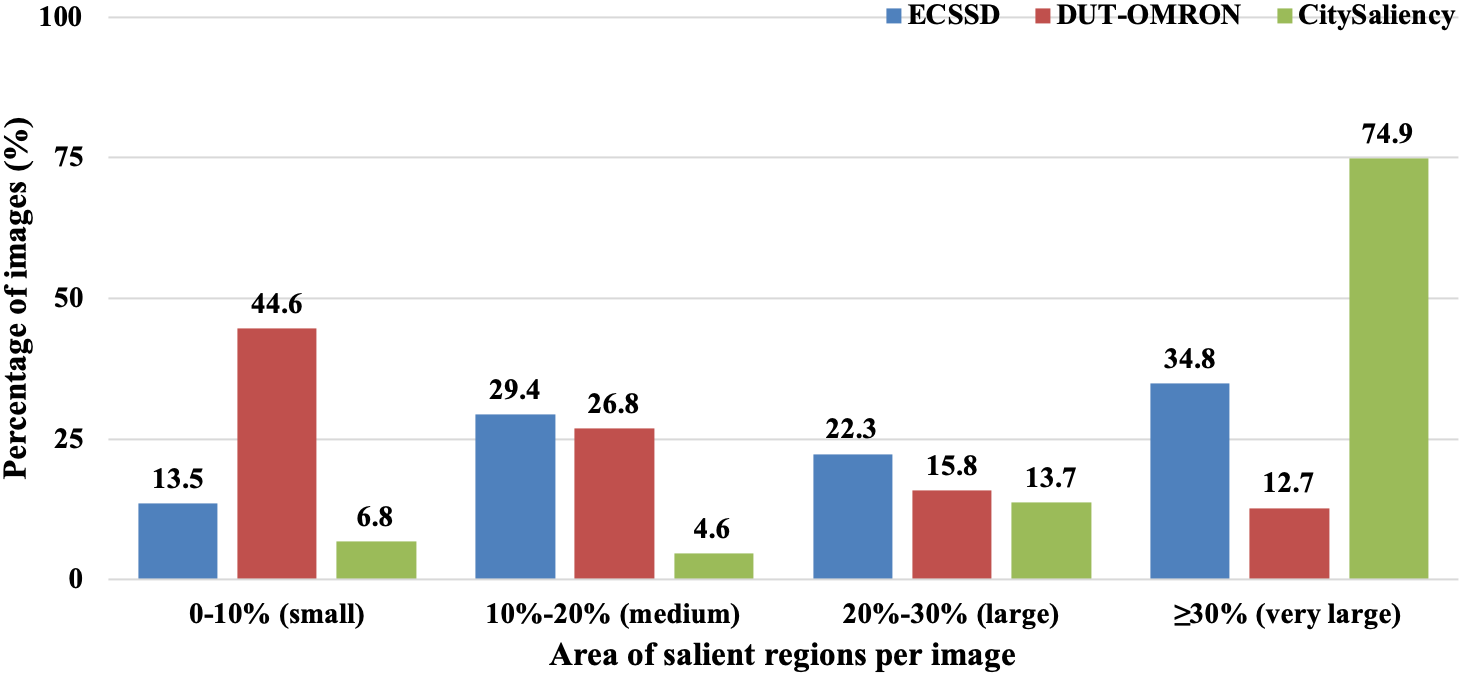}
\vspace{-4mm}
\caption{Histograms of the area of salient objects.}
\label{fig:dataset_area}
\end{figure}

Finally, we obtain CitySaliency with fixation annotation and pixel-wise instance-level salient object annotation. According to the division of Cityscapes, corresponding 2,975 images are for training and 500 images are for testing. For the task of SOD, we make salient objects with same saliency scores \ie, 1 to generate binary masks since we don't need to distinguish different instances. In this paper, we only use the binary masks of CitySaliency except for special instructions.

\begin{figure*}[t]
\centering
\includegraphics[width=1\textwidth, height=6cm]{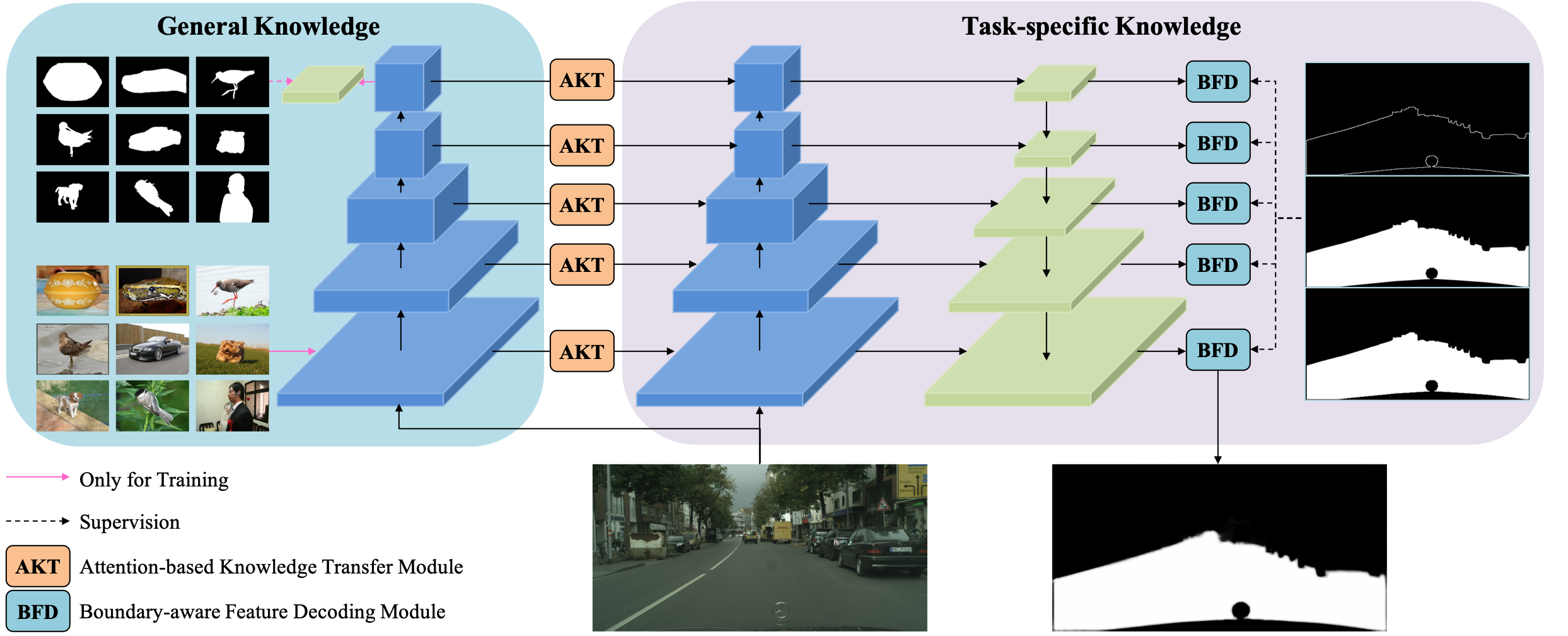}
\vspace{-7mm}
\caption{The framework of the baseline. We first extract the general knowledge by common SOD methods, and then the extracted knowledge is transferred to the task-specific knowledge by an attention-based knowledge transfer module (AKT) to deal with the cross-domain knowledge difference. After that, the task-specific knowledge is decoded by a boundary-aware feature decoding module (BFD) in a progressive manner to detect task-specific salient objects.}
\label{fig:framework}
\end{figure*} 

\subsection{Analysis}
Comparing CitySaliency with conventional datasets, we can find there usually exist different task-related targets as displayed in Fig.~\ref{fig:dataset_motivation}. In addition to the respective of tasks, 
we provide comparisons for the perspective of scenes by presenting the annotation maps (AAMs) of two conventional image-based SOD datasets (ECSSD~\cite{yan2013hierarchical} and DUT-OMRON~\cite{yang2013saliency}) and CitySaliency, as shown in Fig.~\ref{fig:dataset_aam}, which reflects the distribution of salient objects in the dataset. From Fig.~\ref{fig:dataset_aam}, we can see that  conventional datasets are usually center-biased, while the CitySaliency is more discrete. In addition, the histograms of number and area of salient objects are displayed in Fig.~\ref{fig:dataset_number} and~\ref{fig:dataset_area}. We can observe that there are usually more salient objects and there is more area occupation in CitySaliency. 

From these observations, we can find two main difficulties \ie, the cross-domain knowledge difference and task-specific scene gap (including more discrete distribution, more salient objects and more area occupation), which makes it difficult to deal with the task-aware SOD for conventional SOD methods.


\section{A Baseline Model}
To address these difficulties (\ie, the cross-domain knowledge difference and task-specific scene gap), we propose a baseline model for driving task-aware SOD via a knowledge transfer network. The framework of our approach is shown in Fig.~\ref{fig:framework}. 

\subsection{General Knowledge Extraction}
To extract the general knowledge, we construct the general subnetwork taking ResNet-50~\cite{he2016deep} with feature pyramid as the feature extractor, which is modified by removing the last two layers for the pixel-level prediction task. As shown in Fig.~\ref{fig:framework}, the feature extractor has five residual stages denoted as $\mathcal{E}_\mathcal{G}^{(i)} (\pi_\mathcal{G}^{(i)}), i =\{1, \dots, 5\}$, where $\pi_\mathcal{G}^{(i)}$ is the set of parameters of $\mathcal{E}_\mathcal{G}^{(i)}$. In addition, to obtain larger feature maps and receptive fields, we set the strides of last residual stages $\mathcal{E}_\mathcal{G}^{(5)}$ to 1 and dilation rates of last two stages to 2 and 4. The subnetwork is trained by the conventional datasets and learns the general knowledge (\ie, the output of $\mathcal{E}_\mathcal{G}^{(i)}$).

\subsection{Attention-based Knowledge Transfer}
To deal with the knowledge difference between conventional datasets and CitySaliency, we propose an attention-based knowledge transfer module (AKT), which transfers the extracted general features in general subnetwork to task-specific features in task-specific subnetwork. 

\begin{figure}[t]
\centering
\vspace{-5mm}
\includegraphics[width=1.00\columnwidth,height=2.2cm]{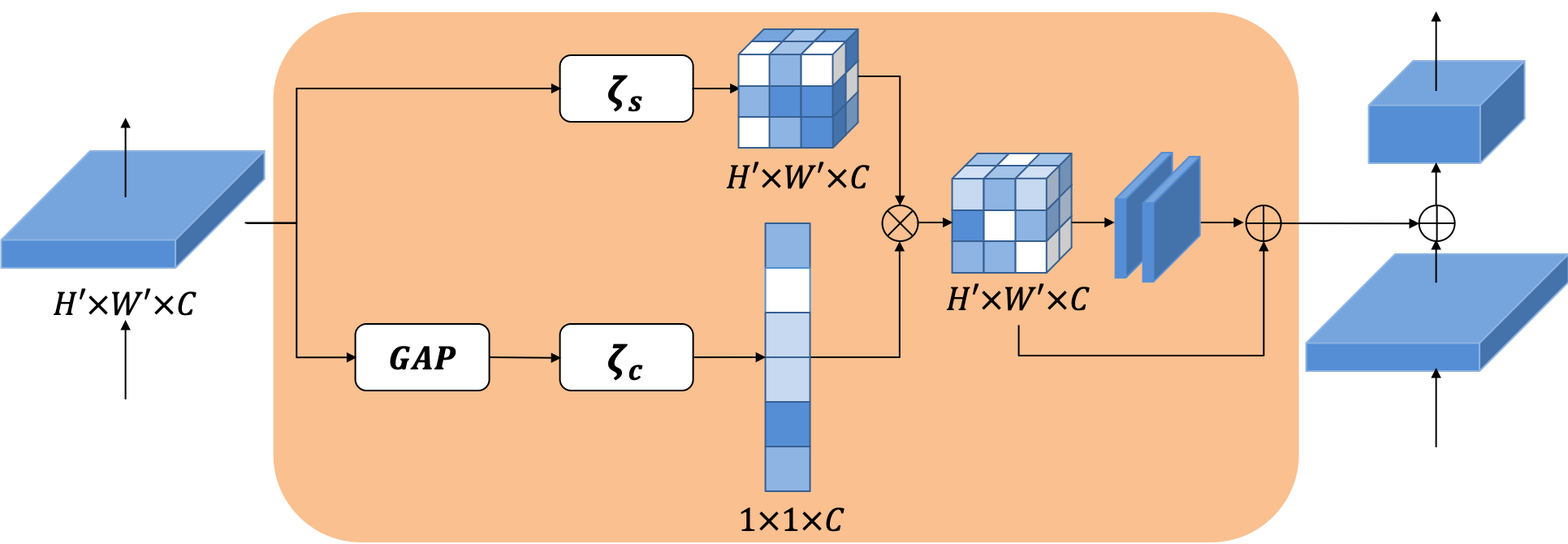}
\vspace{-8mm}
\caption{The structure of AKT. }
\label{fig:model_akt}
\end{figure}

As shown in Fig.~\ref{fig:model_akt}, AKT is based on attention mechanism to purify general knowledge. For the sake of simplification, we denote the input convolutional features as $\text{F}_\mathcal{G} \in \mathbb{R}^{H' \times W' \times C}$, which is the output of $\mathcal{E}_\mathcal{G}^{(i)}, i = {1, \dots, 5}$. To purify the features, we can compute an attention map as follows:
\begin{equation}
\text{A}_{{\mathcal{P}}} = \zeta_{s}(\text{F}_{{\mathcal{G}}}) \otimes \zeta_{c} (GAP(\text{F}_{{\mathcal{G}}})),
\label{eq:promotion_attention}
\end{equation}
where $\zeta_{s}(\cdot)$ and $\zeta_{c}(\cdot)$ denotes the Softmax operation on the spatial and channel dimension respectively, $GAP(\cdot)$ means global average pooling, and $\otimes$ and $\oplus$ represents element-wise product and summation. Based on the attention map $\text{A}_{{\mathcal{P}}}$ from general knowledge, we use a residual module to learn the cross-domain knowledge difference from the general domain to task domain and the learned shift is added to the task-specific features $\text{F}_\mathcal{T} \in \mathbb{R}^{H' \times W' \times C}$ produced by task-specific subnetwork. As shown in Fig.~\ref{fig:framework},   the task-specific subnetwork has the same feature extractor as general subnetwork and its residual stages are denoted as $\mathcal{E}_\mathcal{S}^{(i)} (\pi_\mathcal{S}^{(i)}), i =\{1, \dots, 5\}$, where $\pi_\mathcal{S}^{(i)}$ is the set of parameters of $\mathcal{E}_\mathcal{S}^{(i)}$. 

For knowledge transfer, we put AKT after the output of $\mathcal{E}_\mathcal{G}^{(i)}$ to learn the knowledge difference, which are added to $\mathcal{E}_\mathcal{S}^{(i)}$ as the input of $\mathcal{E}_\mathcal{S}^{(i+1)}, \forall i=1, \dots, 4$. For the convenient representation, we denote the AKT behind $\mathcal{E}_\mathcal{G}^{(i)}$ as $\phi_\mathcal{A}^{(i)}(\pi_\mathcal{A}^{(i)})$, where $\pi_\mathcal{A}^{(i)}$ is the set of  parameters of $\phi_\mathcal{A}^{(i)}$. In this manner, the knowledge difference of conventional datasets and CitySaliency can be learned and used to make up for task-aware SOD, assisting to find salient objects.

\subsection{Boundary-aware Feature Decoding}
To deal with the complex scenes in the task domain, we introduce an effective boundary-ware feature decoding module (BFD) to fine decode features of lots of salient objects and their structures. Inspired by~\cite{su2019selectivity}, we make the decoding process pay attention to the boundaries and interiors to fine decode the features from the feature extractor of task-specific subnetwork. During decoding, the features from feature extractor $\mathcal{E}_\mathcal{T}^{(i)} (\pi_\mathcal{T}^{(i)})$ and the feature from the finer decoding layer are integrated to form the features, named as $\text{F}_{\mathcal{D}} \in \mathbb{R}^{H'' \times W'' \times C'}$, which is the input of BFD.

\begin{figure}[t]
\centering
\vspace{-5mm}
\includegraphics[width=1.00\columnwidth,height=2.2cm]{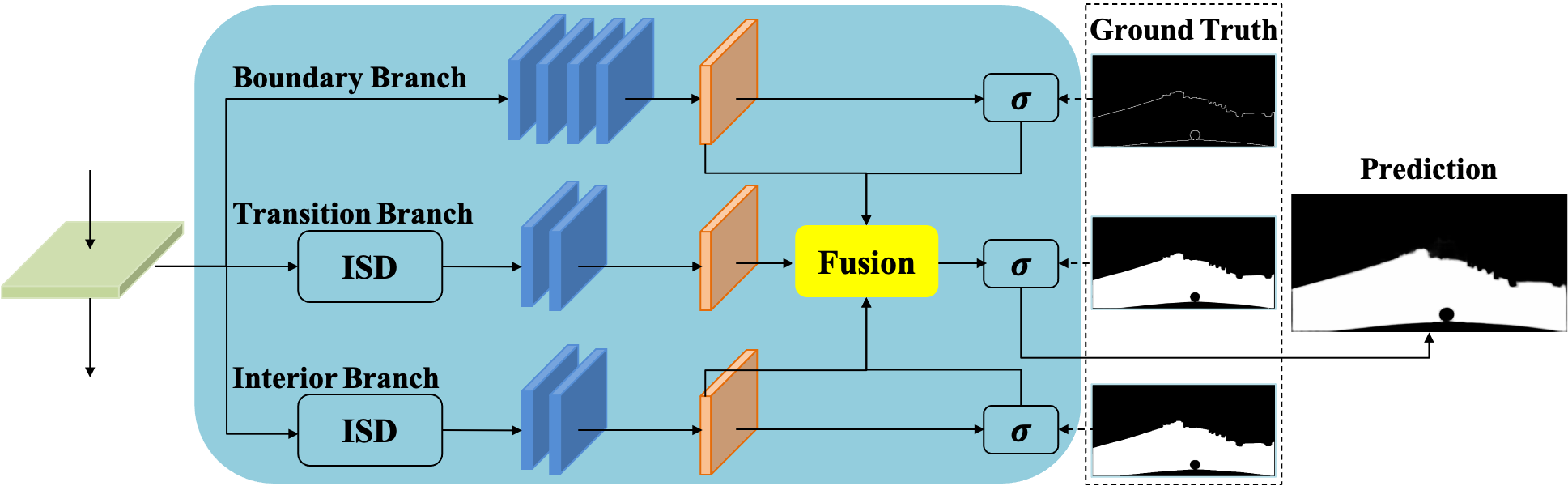}
\vspace{-8mm}
\caption{The structure of BFD.}
\label{fig:model_bfd}
\end{figure}

As shown in Fig.~\ref{fig:model_bfd}, BFD has three parallel branches, \ie, boundary branch ($\phi_\mathcal{B}(\pi_\mathcal{B})$), transition branch ($\phi_\mathcal{T}(\pi_\mathcal{T})$) and interior branch ($\phi_\mathcal{I}(\pi_\mathcal{I})$), where $\pi_\mathcal{B}$, $\pi_\mathcal{T}$ and $\pi_\mathcal{I}$ are the set of parameters of three branches, respectively. In the boundary branch, two convolution and one transposed  layers are stacked. In the transition branch, an integrated successive dilation module from~\cite{su2019selectivity} is used to integrate various scale context. The interior branch has the similar structure with the transition branch. Let $\textbf{F}_\mathcal{T}$ be the fused features, which is obtained by the fusion of three maps according to the boundary confidence map $\textbf{M}_\mathcal{B} = \sigma(\phi_\mathcal{B})$ and the interior confidence map $\textbf{M}_\mathcal{I} = \sigma(\phi_\mathcal{I})$.  Finally, we obtain the prediction of salient objects as $ \textbf{M}_0 = \sigma(\textbf{F}_\mathcal{T})$.

We put the BFD after the each decoding stage of task-specific subnetwork as shown in Fig.~\ref{fig:framework}. For the sake of simplification, we add the index for each confidence maps as $\textbf{M}_\mathcal{B}^{(i)}$, $\textbf{M}_\mathcal{T}^{(i)}$ and $\textbf{M}_\mathcal{I}^{(i)}, i = {1, \dots, 5}$. Therefore, the overall learning objective can be formulated as:
\begin{equation}
\begin{split}
\min_{\mathbb{P}} \sum_{i =1}^5 L(\textbf{M}_0^{(i)}, G_0) + L(\textbf{M}^{(i)}_{B}, G_\mathcal{B}) + L(\textbf{M}^{(i)}_{I}, G_\mathcal{I}),
\end{split}
\label{eq:overall_loss}
\end{equation}
where $\mathbb{P}$ is the set of $\{\pi^{(i)}_{\mathcal{A}}, \pi^{(i)}_{\mathcal{S}}, \pi^{(i)}_{\mathcal{B}}, \pi^{(i)}_{\mathcal{T}}, \pi^{(i)}_{\mathcal{I}}\}^5_{i=1}$ for convenience of presentation, $L(\cdot, \cdot)$ is the binary cross-entropy loss, and $G_0, G_{\mathcal{B}}, G_{\mathcal{I}}$ represents the ground-truth masks of salient objects, boundaries and interiors, respectively.


\section{Experiments}
\subsection{Settings}
\subsubsection{Evaluation Metrics}
We choose mean absolute error (MAE), weighted F-measure score ($F^w_{\beta}$), F-measure score ($F_{\beta}$) and S-measure ($S_m$) to evaluate methods on CitySaliency. 
\subsubsection{Training and Inference}

We use stochastic gradient descent algorithm to train our network. The network is trained in two stages as follow. 1) We first train the general subnetwork. In the optimization process, parameters of the feature extractor are initialized by pre-trained ResNet-50 model with learning rate of $1 \times 10^{-3}$, weight decay of $5 \times 10^{-4}$ and momentum of 0.9. And learning rates of rest layers are set to 10 times larger. The general subnetwork is trained on DUTS~\cite{wang2017stagewise} to learn general knowledge. Training images are resized to the resolution of $512 \times 256$, and applied horizontal flipping. The training process takes 50,000 iterations with mini-batch of 4. 2) We fix the general subnetwork to train the rest network (including AKT and task-specific subnetwork) with the same setting as general subnetwork and the optimization objective is Eq.~(\ref{eq:overall_loss}). We use CitySaliency to training the network. The training process takes 200,000 iterations. During inference, one image is directly fed into the network to produce the saliency map at the output of first stage in the task-specific subnetwork.

\begin{figure}[t]
\centering
\includegraphics[width=1\columnwidth, height=2.8cm]{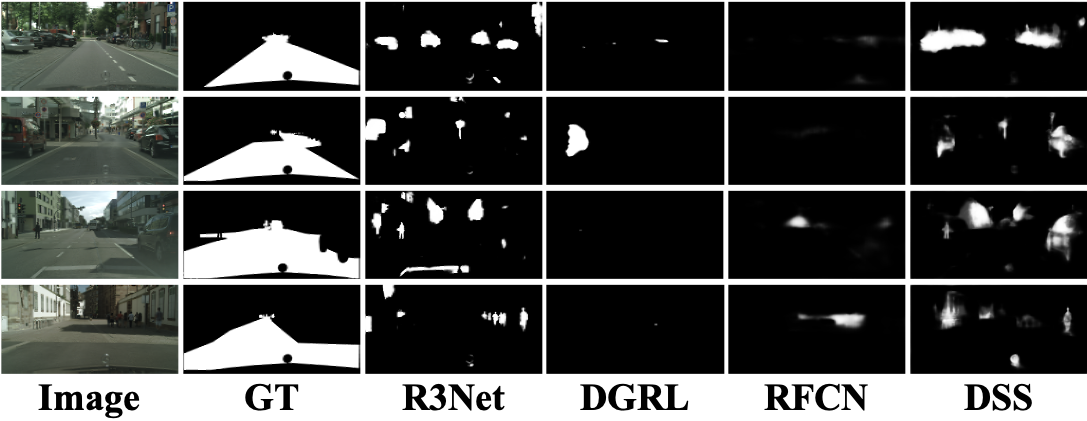}
\vspace{-7mm}
\caption{Examples of the state-of-the-art methods on CitySaliency without fine-tuning. GT means ground truth.}
\label{fig:result_before}
\end{figure}

\subsection{Model Benchmarking on CitySaliency}
To analyze the proposed dataset, we first benchmark 12 state-of-the-art model performance in Tab.~\ref{tab:performance_before} before fine-tuning on CitySaliency dataset. Some examples are shown in Fig.~\ref{fig:result_before}. From the quantitative and qualitative evaluation, we can find that it is difficult for these models to find correct salient objects and to restore their fine structures.

\begin{figure}[t]
\centering
\includegraphics[width=1\columnwidth]{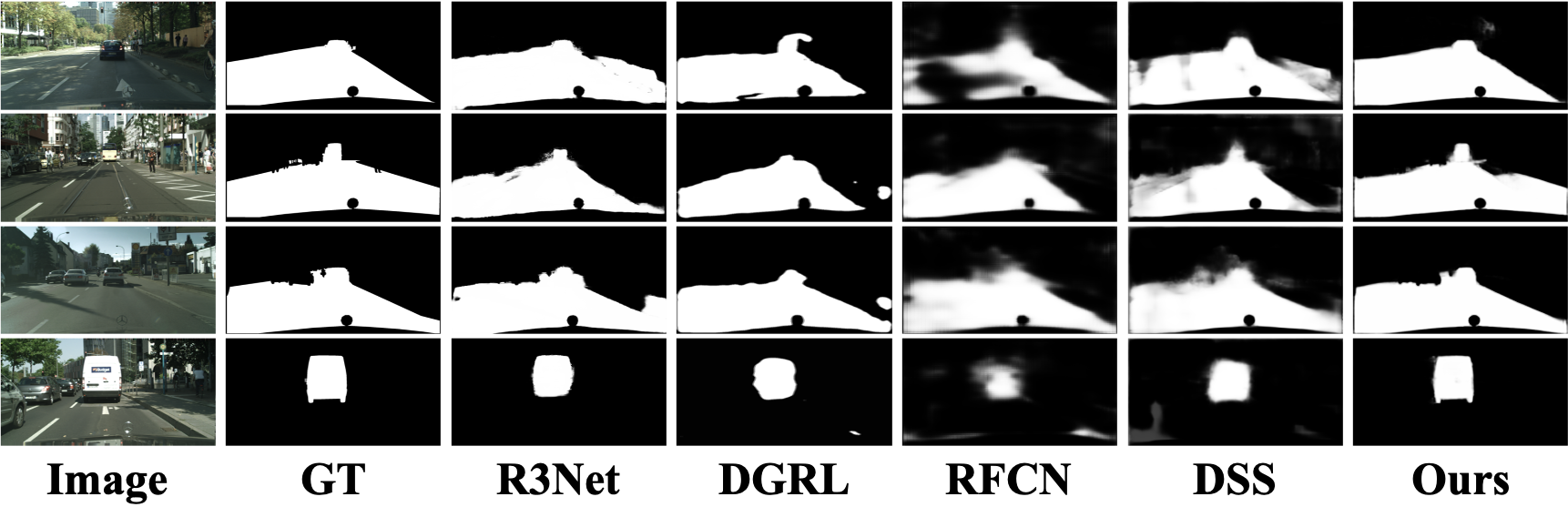}
\vspace{-7mm}
\caption{Representative examples of the state-of-the-art methods and our approach after being fine-tuned.}
\label{fig:result_after}
\end{figure}

Beyond the comparisons without fine-tuning, we fine-tune these models and the proposed method on CitySaliency. The performance comparisons of these models are listed in Tab.~\ref{tab:performance_before} and shown in Fig.~\ref{fig:result_after}. we can find that the performance of these methods is all improved, which validates the knowledge difference of conventional datasets and CitySaliency. Also, our method consistently outperforms other state-of-the-art methods on four metrics, even if the dense CRF is used on R3Net and DSS. We can also find the proposed method has better locations and structures of salient objects. 

From these experiments, we can believe that task-specific CitySaliency is a challenging dataset, and the proposed model with knowledge transfer and feature decoding are useful for solving the task-aware SOD.

\begin{table}[t]
\centering
\vspace{-4.5mm}
\small
\caption{Performance of 13 state-of-the-art models before and after fine-tuning on CitySaliency. The best two results are in \textbf{\color{red}{red}} and \textbf{\color{blue}{blue}}. ``$^\dagger$'' means utilizing dense CRF~\cite{hou2019deeply}.}
\setlength{\tabcolsep}{0.5mm}{
\renewcommand\arraystretch{1.0}
\begin{tabular}{c | c c c c | c c c c}
\hline
& Before & & & & After \\
\hline
 & MAE 
 & $F^w_\beta$ 
 & $F_\beta$
  & $S_m$
   & MAE
 & $F^w_\beta$ 
 & $F_\beta$ 
  & $S_m$
 \\
\hline
UCF\cite{zhang2017learning} & 0.428 & \textbf{\color{red}{0.208}} & 0.291 & \textbf{\color{red}{0.361}} & 0.344 & 0.468 & 0.515 & 0.583 \\
NLDF\cite{luo2017non} & 0.369 & 0.084 & 0.226 & 0.332 & 0.172 & 0.676 & 0.749 & 0.744   \\
Amulet\cite{zhang2017amulet} & 0.416 & \textbf{\color{blue}{0.175}} & 0.274 & 0.333 & 0.170 & 0.671 & 0.766 & 0.737 \\
FSN\cite{chen2017look} & 0.365 & 0.085 & 0.230 &0.339 & 0.146 & 0.725 & 0.780 & 0.755 \\
SRM\cite{wang2017stagewise} & \textbf{\color{red}{0.358}} & 0.099 & \textbf{\color{blue}{0.299}} & \textbf{\color{blue}{0.349}} & 0.171 & 0.639 & 0.759 & 0.721 \\
RAS\cite{chen2018eccv} & 0.380 & 0.075 & 0.174 &0.314 & 0.147 & 0.716 & 0.797 & \textbf{\color{blue}{0.771}} \\
PiCANet\cite{liu2018picanet} & 0.372 & 0.112 & 0.259 & 0.340 & 0.189 & 0.656 & 0.768 & 0.749 \\
R3Net$^\dagger$\cite{deng2018r3net} & 0.392 & 0.074 & 0.153 &0.292 & \textbf{\color{blue}{0.140}} & \textbf{\color{blue}{0.741}} & \textbf{\color{blue}{0.805}} & 0.760 \\
DGRL\cite{wang2018detect} & \textbf{\color{blue}{0.360}} & 0.086 & \textbf{\color{red}{0.401}} & 0.342 & 0.146 & 0.701 & 0.725 & 0.725\\
RFCN\cite{wang2018salient} & 0.364 & 0.090 & 0.229 &0.342 & 0.168 & 0.672 & 0.766 & 0.747 \\
DSS$^\dagger$\cite{hou2019deeply} & 0.386 & 0.080 & 0.185 &0.315 & 0.155 & 0.697 & 0.790 & 0.767 \\
BANet\cite{su2019selectivity} & 0.375 & 0.112 & 0.258 & 0.331 & 0.144 & 0.735 & \textbf{\color{blue}{0.805}} & \textbf{\color{blue}{0.771}} \\
\hline
\textbf{Ours} & - & - & - & - & \textbf{\color{red}{0.133}} & \textbf{\color{red}{0.751}} & \textbf{\color{red}{0.811}} & \textbf{\color{red}{0.785}} \\
\hline
\end{tabular}}
\label{tab:performance_before}
\end{table}

\subsection{Performance Analysis of the Baseline Model}
To investigate the effectiveness of the proposed domain-aware network, we conduct ablation experiments by introducing four different models. The first setting is only the task-specific subnetwork without AKT and BFD, which is regraded as ``Baseline''. To explore the effectiveness of AKT, we add the second (named as ``Baseline + PT'', pre-trained on DUTS-TR and then fine-tuned on CitySaliency) and third (``Baseline + AKT'', with AKT) models. In addition, we add the fourth model arming ``Baseline'' with BFD as ``Baseline + BFD''. We also list the proposed model as ``\textbf{Ours}''.

The comparisons of above models are listed in Tab.~\ref{tab:performance_ablation}. By comparing first three rows, we can observe the pre-trained operation on conventional datasets is useful for task-aware SOD, and AKT is more effective to utilize the general knowledge from conventional datasets. In addition, we can find ``Baseline + BFD'' also has an obvious improvement compared with ``Baseline'', which indicates the effectiveness of BFD. Moreover, the combination of the attention-based knowledge transfer and boundary-aware feature decoding achieve a better performance, which verifies the usefulness of the proposed baseline model. As a conclusion, this model verifies the feasibility of solving driving-aware SOD and push the future research.

\begin{table}[t]
\centering
\vspace{-6mm}
\small
\caption{Performance of ablation experiments.}
\setlength{\tabcolsep}{3mm}{
\renewcommand\arraystretch{1.0}
\begin{tabular}{c | c c c c}
\hline
 & MAE $\downarrow$
 & $F^w_\beta$ $\uparrow$
 & $F_\beta$ $\uparrow$
  & $S_m$ $\uparrow$
 \\
\hline
Baseline & 0.149 & 0.721 & 0.796 & 0.770 \\
Baseline + PT & 0.143 & 0.726 & 0.797 & 0.772 \\
Baseline + AKT & 0.141 & 0.735 & 0.810  & 0.782  \\
Baseline + BFD & 0.140 & 0.737 & 0.810 & 0.780 \\
\hline
\textbf{Ours} & \textbf{0.133} & \textbf{0.751} & \textbf{0.811} & \textbf{0.785} \\
\hline
\end{tabular}}
\label{tab:performance_ablation}
\end{table}

\section{Conclusion}

In this paper, we construct a new driving task-specific salient object detection (SOD) dataset in urban driving to explore SOD in special tasks. By analyzing this dataset, we find two main difficulties, which prevent the development of task-aware SOD. Inspired by the difficulties, a baseline model via the knowledge transfer network is proposed. This model is organized in a progressive manner, utilizing the attention-based knowledge transfer module and boundary-aware feature decoding module to deal with the existing difficulties. In addition, we provide a comprehensive benchmark of the state-of-the-art methods and our baseline model on the proposed dataset, which shows the challenges of the task-specific dataset, and validates the effectiveness of the proposed model. This work pays more attention to the salient features of objects in the driving scene and does not focus on the dynamic attribute analysis, which will be explored more with the help of time domain signals in the feature.

\section{Acknowledgement}
This work is partially supported by the National Natural Science Foundation of China under the Grant 61922006, and Baidu academic collaboration program.

\bibliographystyle{IEEEbib}
\bibliography{RefTSOD}

\end{document}